\documentclass{article}
\usepackage{iclr2027_conference,times}
\usepackage[T1]{fontenc}
\usepackage{amsmath,amssymb,mathtools}
\usepackage{graphicx,booktabs,array}
\usepackage{placeins}
\usepackage{enumitem}
\usepackage{xcolor}
\usepackage{microtype,xcolor,algorithm,algpseudocode}
\usepackage{hyperref,url}
\hypersetup{hidelinks}

\newcommand{\ind}{\mathbb{I}}

\DeclareMathOperator{\LN}{LN}

\title{DRelay: Global Draft Context for Prefix-Aware Parallel Speculative Decoding Repair}
\author{
Zhuoyu Wang \qquad Junnan Huang \qquad Xinyu Chen$^{\dagger}$\\[3pt]
\normalfont The Hong Kong University of Science and Technology (Guangzhou)\\[3pt]
\normalfont{\small $^{\dagger}$Corresponding author.}
}
\iclrfinalcopy
\begin{document}
\maketitle
\fancyhead{}
\begin{abstract}
Parallel drafting reduces the drafting overhead of speculative decoding for large language models (LLMs), but its gains remain limited by the accepted prefix length. Even when the correct token is present in the candidate pool, a single early selection error prevents subsequent predictions from being used. We propose DRelay, which uses global information from the entire draft block to perform prefix-aware selective repair of candidate selections before target-model verification. DRelay bases its decisions on both candidate correlations and the selected path: a global reader extracts predictive information across positions for each candidate. While a causal selector combines candidate-level information extracted by the global read with the tokens selected at preceding positions to determine whether the native choice at the current position is consistent with the global evidence and the selected prefix. It then decides whether to retain or replace the token, thereby repairing early errors and extending the accepted prefix. We further jointly train the draft backbone and the selector, combining candidate-support learning with a repair objective, while weighting the repair loss according to each block position’s potential contribution to the consecutive accepted prefix. Across eight diverse benchmarks on an H800 GPU, DRelay consistently improves both average acceptance length and end-to-end decoding performance over DFlash, Domino, and DSpark. Under SGLang serving, DRelay improves average end-to-end speedup over DFlash, Domino, and DSpark by $14.7\%$--$16.8\%$, $8.7\%$--$9.3\%$, and $8.1\%$--$9.3\%$, respectively. 

\end{abstract}
\section{Introduction}
Speculative decoding reduces the overhead of token-by-token autoregressive inference in large language models through inexpensive drafting and batched verification. Parallel drafting further reduces drafting overhead by predicting multiple positions in a single forward pass \citep{leviathan2023speculative,chen2023speculative} . DFlash \citep{chen2026dflash} adopts this approach to predict an entire masked draft block in parallel within a single forward pass. Building on this approach, Domino and DSpark \citep{huang2026domino,cheng2026dspark} model dependencies among draft tokens through a GRU state and transitions conditioned on preceding tokens, respectively, allowing subsequent predictions to condition on the selected prefix and thereby increasing acceptance length.

However, speculative decoding imposes a key constraint: each verification round can accept only a contiguous correct prefix starting at the first position of the draft block. Consequently, an error at an early position prevents subsequent tokens from being committed in the current round, even if those tokens have been predicted correctly. Moreover, when subsequent predictions are conditioned on an incorrect token, their conditioning context deviates from the correct prefix, so the early error may further affect suffix predictions. This creates an inherent limitation for methods such as Domino and DSpark that introduce additional autoregressive modules: they rely on forward causal dependencies to improve subsequent choices, but information in later predictions cannot directly revise choices already made at earlier positions.

In the base draft model, the correct token is often already present in the candidate set when an early prediction error occurs. Correcting an early selection can therefore enable acceptance of the current token and potentially a subsequent run of correctly predicted tokens. An underutilized opportunity lies in backward information flow within the draft block: global information can guide candidate selection at earlier positions. After candidate generation, we focus on explicitly revisiting the distributions produced by that forward pass and using global information to repair early selections.

Our central idea is for the entire draft block to serve two roles simultaneously: proposing candidates at each position and providing context for candidate selection. Because predictions at later positions are already available before verification, we can identify the most suitable candidate at an earlier position by comparing candidates’ compatibility with global information from the entire draft block and with subsequent predictions. Once a candidate is selected, that specific token should condition subsequent decisions, allowing later choices to follow the actual selected or repaired prefix.

\begin{figure}[t]
\centering
\includegraphics[width=\linewidth]{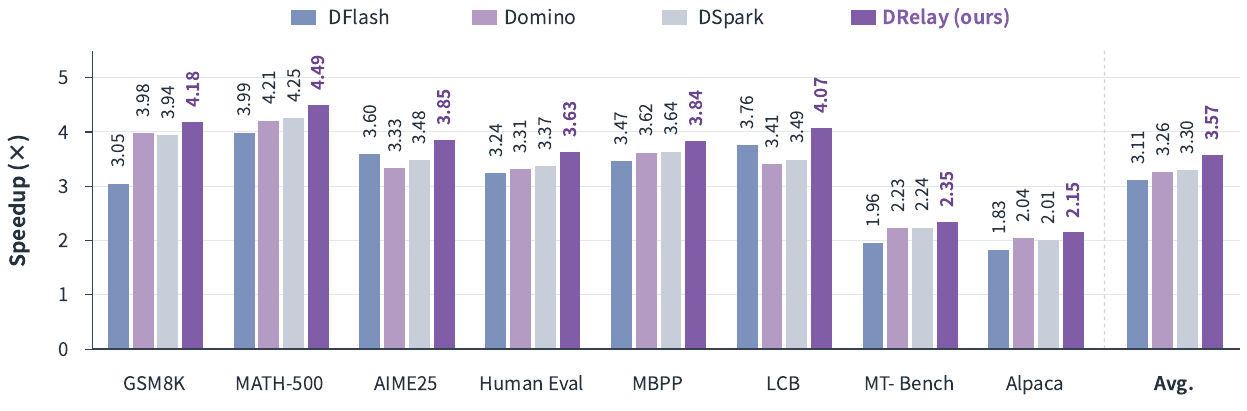}
\caption{End-to-end decoding speedup on Qwen3-4B. DFlash \citep{chen2026dflash}, Domino \citep{huang2026domino}, DSpark \citep{cheng2026dspark}, and DRelay are compared against autoregressive decoding.}
\label{fig:evaluation}
\end{figure}

We introduce DRelay, which implements this candidate repair process through the following key components and techniques: (1) A global reader first integrates candidate distributions across positions and then performs reads across positions for each candidate to construct a global context tailored to each alternative token. (2) A causal selector combines global context, the verified target-model history, and an addressable representation of the selected prefix to sequentially decide whether to retain the original draft token or select an alternative candidate. (3) We jointly train the backbone and the selector to learn candidate distributions and repair actions, respectively, and use these two factors to weight the repair loss, focusing training on choices that contribute more to the accepted prefix. As shown in Figure~\ref{fig:evaluation}, DRelay outperforms DFlash, Domino, and DSpark across all displayed benchmarks on Qwen3-4B, achieving up to a 4.49× end-to-end speedup over autoregressive decoding. Our main contributions are:

\begin{itemize}\setlength{\itemsep}{1pt}
\item We propose DRelay, which uses the entire draft block to guide repairs to early selections and allows subsequent candidates to directly query the actual repaired prefix, completing candidate selection sequentially before verification.
\item We design a joint training objective that simultaneously trains the draft model to propose correct candidates and the selector to make repair decisions, while weighting the repair loss according to prefix value.
\item We evaluate DRelay on Qwen3-4B and Qwen3-8B across eight math, code, dialogue, and instruction-following benchmarks. Across target models and temperatures, its average end-to-end speedup exceeds that of the best-performing baseline among DFlash, Domino, and DSpark by 7.8--9.9\%, while concurrent-serving throughput reaches up to 4.38× that of autoregressive decoding at matched concurrency.
\end{itemize}

\section{Related Work}
\paragraph{Parallel drafting and causal selection.}
Speculative decoding amortizes target computation by verifying several
proposals together \citep{leviathan2023speculative,chen2023speculative}.
DFlash obtains parallel proposals through noncausal attention over
target-conditioned masked slots \citep{chen2026dflash}; DFlare improves how
deeper draft layers use target features \citep{zhang2026dflare}.
Domino restores within-block dependence with a GRU-conditioned vocabulary
residual \citep{huang2026domino}. DSpark uses a preceding-token Markov head
by default, offers a recurrent variant, and schedules verification length
according to confidence and load \citep{cheng2026dspark}.
These works establish the parallel-backbone/causal-head split. DRelay adds
an explicit read of the resulting candidate distributions, then conditions
each decision on addressable selected-prefix entries. The distinction is
where selection obtains evidence, rather than whether the backbone has
noncausal attention.

\paragraph{Draft-model training and distillation.}
Target-aligned drafters are trained with supervised or knowledge-distillation
objectives. DistillSpec uses on-policy trajectories and a decoding-aware
divergence \citep{zhou2024distillspec}; EAGLE and EAGLE-3 learn target
features through autoregressive feature prediction and training-time test
\citep{li2024eagle,li2025eagle3}. For parallel drafters, DFlash predicts a
masked block in one pass \citep{chen2026dflash}. Draft-OPD replays
verification-exposed errors under target-assisted rollouts, while D-PACE
derives position weights from a differentiable accepted-length surrogate
\citep{lei2026draftopd,wu2026dpace}. DRelay jointly trains candidate support
and a causal repair selector, weighting repair by prefix value.

\paragraph{Verifier feedback and prefix-conditioned repair.}
TAPS scores path reach to select a compact prefix-closed verification tree
\citep{wang2026taps}. Verification-Aware Training predicts cumulative
acceptance and begins a decaying loss schedule at the first rejection
\citep{gu2026vat}. AdaFlash and online draft co-training adapt drafters to
target-generated data \citep{qian2026adaflash,wang2026onlinecotraining}.
ReTrace and DFlow reuse rejected-trajectory or verifier information across
rounds \citep{lin2026retrace,zhang2026dflow}. DRelay instead reads predicted
evidence within the current block, alongside strictly verified target
history, and uses that evidence to make addressable repair decisions before
verification.

\section{Losses from Candidate Selection and Repair Opportunities in Parallel Drafting}
\label{sec:motivation}

Parallel drafting generates an entire block of predictions at once, but verification can accept only a contiguous prefix starting at the beginning of the block. We investigate an opportunity enabled by parallel drafting: once the entire block has been predicted, can its candidate distributions and content at later positions help correct early selections that truncate the accepted prefix? Using native greedy decoding traces from DFlash and Domino with Qwen3-4B \citep{yang2025qwen3} as the target model, we analyze this opportunity from three perspectives: candidate availability, information from later predictions, and gains in accepted prefix length.

\begin{figure}[!ht]
\centering
\includegraphics[width=\linewidth]{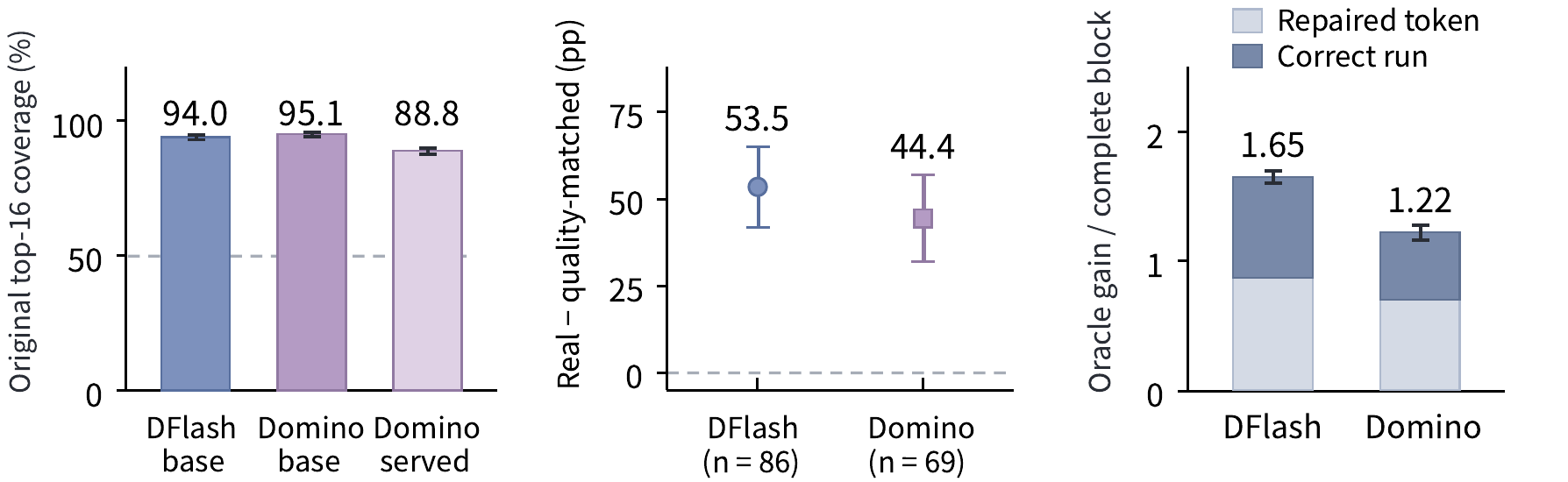}
\caption{\textbf{Candidate availability, continuation evidence, and repair value.}
\textbf{Left:} original top-16 coverage at first base/served errors
($n=7{,}787/5{,}348/4{,}825$).
\textbf{Middle:} real-minus-correctness-matched best-alternative accuracy at
covered early served errors ($n=86/69$), using served pools (post-head for Domino). Receiver and donor each have four correct subsequent predictions,
an oracle condition. Intervals are 97.5\% per model for two primary comparisons.
\textbf{Right:} original-pool one-edit gain over 8,057/5,896 complete blocks,
separating the repaired token from the recovered consecutive run.
Left/right intervals are 95\%; all intervals resample prompts within task.
These panels measure repair opportunities, not trained DRelay performance.}
\label{fig:motivation}
\end{figure}

\subsection{The Correct Candidate Usually Remains Available at the First Error}
An early prediction error does not necessarily imply that the correct candidate is absent. As shown in Figure~\ref{fig:motivation}, at the first error in the base predictions, the original top-16 candidate pool contains the target token in 94.0\% of DFlash cases and 95.1\% of Domino cases.
We separately analyze the first rejection in Domino’s actual draft sequences and find that the original candidate pool still contains the target token in 88.8\% of cases. These results show that suitable alternatives for repair are already available at many error positions.

\subsection{Later Predictions Provide Discriminative Signals for Early Candidate Selection}
Candidate coverage only indicates whether the correct token is present in the
candidate pool; it does not determine which candidate should be selected when
multiple alternatives are available.  We therefore ask whether predictions at
later positions in the draft block can provide additional evidence for an
earlier candidate.  Intuitively, a correct early candidate should not only be
consistent with the prefix at its own position, but should also support a
plausible continuation of the later predictions.

To test this hypothesis, we select samples with an early error for which the
correct token is present in the candidate pool and the next four draft tokens
are all correct.  For each candidate at the current position, we treat that
candidate as the selected token and use a frozen target model to compute the
total conditional log-probability of the four subsequent draft tokens given
the preceding prefix and the candidate.  The candidate with the highest score
is considered the candidate supported by the subsequent predictions.  As a
control, we replace the four subsequent tokens with a suffix from another
prompt, thereby changing the semantics of the sequence, and compare the
resulting token-repair accuracy.  As shown in the middle panel of
Figure~\ref{fig:motivation}, the real suffix selects the correct candidate in
46/86 DFlash cases and 31/69 Domino cases.  Relative to the matched control
suffixes, the accuracy improves by 53.5 and 44.4 percentage points,
respectively.  Thus, subsequent predictions contain context-dependent
discriminative information that can help repair an early error, rather than
merely providing a suffix unrelated to the preceding candidates.

The entire draft block can therefore serve not only as a source of candidates,
but also as global evidence for candidate selection.  DRelay accordingly
combines each candidate's native confidence with global information from the
draft block and the consistency of subsequent predictions.  After a candidate
is selected, its token is fed into the subsequent decisions through the
selected prefix.

\subsection{Early Repair Can Recover Subsequent Correct Predictions}
The value of a repair is not limited to the position that it corrects.  Let
$R$ denote the number of draft tokens immediately following the first error
that are correct.  If all other predictions remain unchanged, correctly
repairing this error increases the accepted prefix length by $1+R$: one token
comes from the repaired position and $R$ tokens come from the subsequently
recovered consecutive predictions.

To quantify this potential gain, we perform a one-edit oracle experiment.  If
the target token at the first rejected position is contained in the original
top-16 candidate pool, we replace the erroneous token with the target token and
leave every other position unchanged.  As shown in the right panel of
Figure~\ref{fig:motivation}, when this gain is averaged over all complete draft
blocks and blocks for which the replacement is unavailable contribute zero,
the mean accepted length increases by 1.65 and 1.22 tokens for DFlash and
Domino, respectively.

This result shows that the value of a repair should be measured by how much
accepted prefix it enables.  DRelay is not restricted to the first error: it
evaluates every position and selectively repairs a position when the global
continuation and the selected prefix indicate that the native draft may be
inconsistent.  Together, the three observations motivate a selector that
reads global information, performs repairs selectively, and propagates each
selected repair to subsequent decisions.  During training, the repair loss
should assign greater weight to decisions according to their
potential contribution to the consecutive accepted prefix, improving the
prefix length realized at inference time.

\section{DRelay: Global Candidate Reads and Causal Selection}
\label{sec:method}

\begin{figure}[t]
\centering
\includegraphics[width=\linewidth]{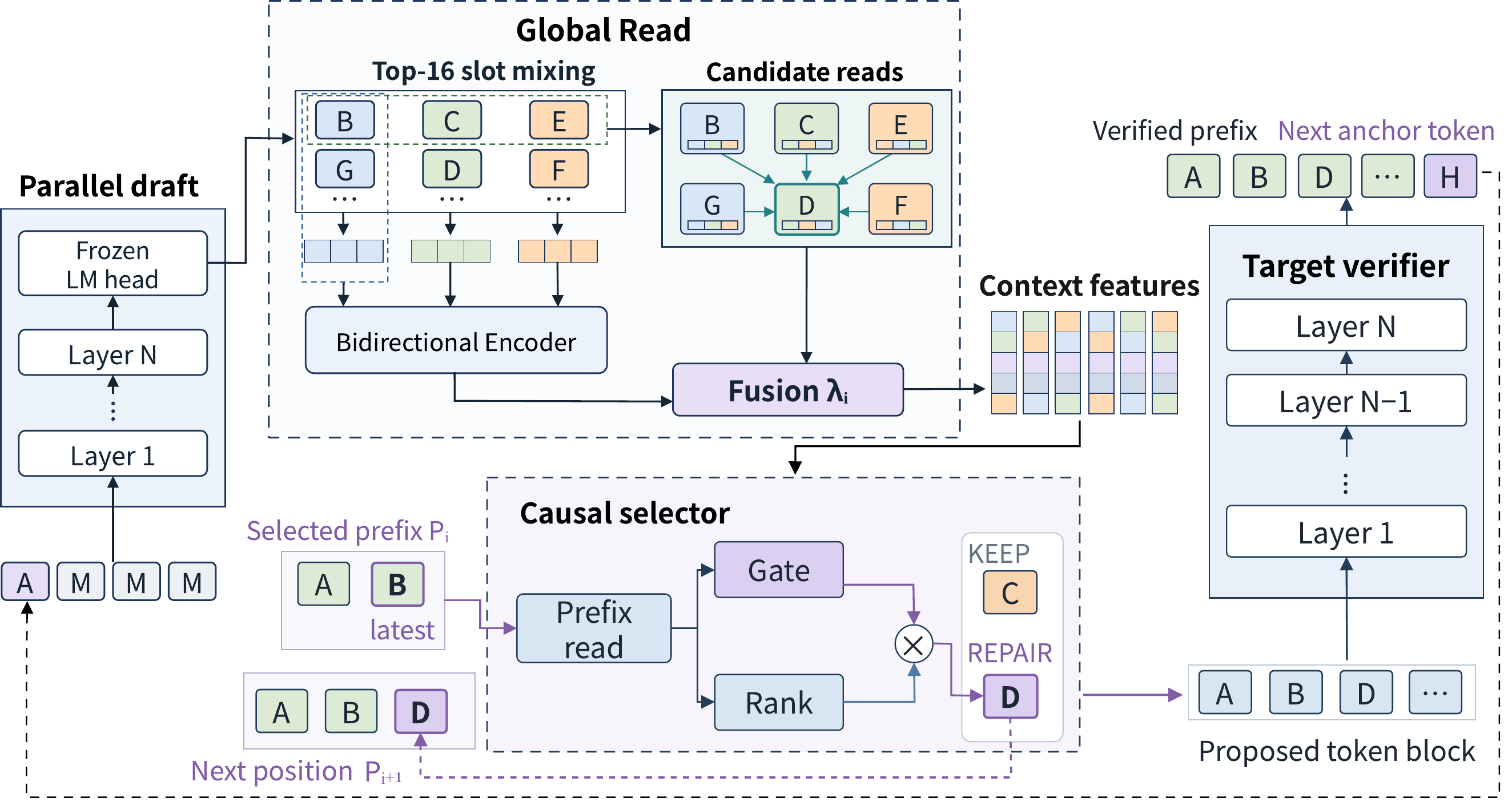}
\caption{\textbf{DRelay reads globally and decides causally.}
The parallel draft predicts candidate pools from anchor A and masked slots M.
Global Read combines a top-16 slot mixture with candidate-specific reads;
Fusion $\lambda_i$ supplies contextual features to the Causal selector.
Prefix read accesses $P_i=[x_0,\ldots,x_{i-1}]$, the selected prefix. Gate and Rank jointly choose
KEEP or REPAIR: replacing C with D extends the prefix to $[\mathrm{A,B,D}]$,
so the next decision uses D. The target verifies the proposed path and supplies
the next anchor. Anchor-feature and verified-history projections are omitted
for clarity.}
\label{fig:architecture}
\end{figure}
DRelay constructs a candidate path after parallel drafting and before target-model verification. As shown in Figure~\ref{fig:architecture}, DRelay organizes path repair into two modules. Global Read first aggregates candidate distributions across positions to construct a global context for each candidate. The Causal Selector then combines this context with the verified history and the selected prefix to decide, at each position, whether to retain the base prediction or choose an alternative candidate, producing a path for target-model verification. 

\subsection{Overall architecture and execution process}
Figure~\ref{fig:architecture} illustrates a single round of DRelay execution. At the start of each round, the parallel drafter receives an anchor token confirmed by the target model and historical context features, then generates hidden states and candidate distributions for subsequent positions in a single forward pass.In Figure~\ref{fig:architecture}, A denotes the anchor token, and M denotes positions to be predicted. The drafter retains the top-16 candidates at each position. The Global Read module then processes all candidates in the block. The Top-16 slot mixing stage first aggregates the candidate distributions across positions, and the Candidate reads stage then retrieves information across positions relevant to each candidate.

The contexts produced by these two stages are fused and passed to the Causal Selector. At each position, the Causal Selector combines the global context, verified history, and currently selected prefix to decide whether to retain the base prediction or choose an alternative candidate. After completing candidate selection for the entire block, DRelay submits the resulting single path to the target model for verification. The target model accepts a contiguous prefix of this path and produces the anchor token for the next round, after which drafting resumes.

\subsection{Global Read: compare candidates with the predicted block}
A single summary can obscure competing continuations. DRelay therefore
reads the block at two resolutions. First, two bidirectional attention--FFN
blocks mix slot features with soft candidate summaries
$\sum_k\pi_{ik}e_{ik}$, where
$\pi_{ik}=q_i(v_{ik})/\sum_lq_i(v_{il})$. This gives each slot a shared
interpretation $u_i$ of the predicted block. Second, candidate-specific
attention preserves individual alternatives. With
$n_{ik}=\LN(u_i+W_ne_{ik})$, each of four attention heads computes
\begin{equation}
 G_{ik}=\sum_{j,l}A_{ik,jl}W_Vn_{jl},\qquad
 A_{ik,jl}\propto\exp\!\left(
 \frac{(W_Qn_{ik})^\top W_Kn_{jl}}{\sqrt{d_g}}
 +\log q_j(v_{jl})+b_{j-i}\right).
 \label{eq:global}
\end{equation}
Here $d_g$ is the head width and $b$ a relative-position bias.
The draft prior weights plausible evidence, while the query lets competing
current candidates attend to different continuations.

Uncertain context should not force an edit. Attention is normalized jointly with an additional zero-valued null
key/value, permitting a content-free read, and a learned slotwise fusion scalar
$\lambda_i$ scales the global slot and candidate residuals. The scalar is
shared across candidates; their reads $G_{ik}$ remain distinct.
This contextual fusion is separate from the later KEEP/REPAIR decision.
Neither is supervised with oracle future correctness.
Global Read uses only the hidden representations and candidate distributions
produced during parallel drafting; it does not read future tokens that have
not been verified by the target model. The noncausal backbone has already
performed an initial information fusion among the masked positions. Global
Read therefore does not establish the first interaction across positions;
instead, it explicitly reads the backbone's output candidate distributions
across positions and extracts global information relevant to each candidate.

\subsection{Causal selector: make each repair part of the context}
After replacing a token, later choices must use the replacement.
An addressable prefix memory stores the anchor and each selected entry,
whose key and value encode $(h_j,z_j^{\mathrm{global}},E(x_j),g_a,j)$.
At position $i$, a candidate can retrieve only entries $j<i$; the latest
chosen token $E(x_{i-1})$ also conditions its query directly.
This exposes earlier choices individually instead of compressing them into
one recurrent state. A separate candidate-specific read of the last 64
strictly verified target features provides direct access to established
context instead of relying only on its compression into draft states. Because this history does not change within the block, all its
reads are computed before selection.

The Rank branch combines native log-probabilities with a shared hidden
residual and compact candidate-specific corrections from global, history,
and prefix evidence. Its scores $S_{ik}$ rank nonbase alternatives.
The Gate branch asks a different question: whether the best available
repair justifies leaving the native choice. This conservative gate is intended
to correct likely error positions while minimizing changes to tokens that are
already consistent with the global context and selected prefix. Together they define
\begin{equation}
 r_{ik}=\operatorname{softmax}_{k>0}(S_{ik}),\qquad
 p_i^{\mathrm{act}}(0)=\sigma(-g_i),\quad p_i^{\mathrm{act}}(k)=\sigma(g_i)r_{ik}\quad(k>0).
 \label{eq:action}
\end{equation}
The best alternative $k^*$ replaces the base only if
$g_i+\log r_{ik^*}>0$, with numerical ties resolved to KEEP.
Thus a positive edit tendency alone is insufficient when alternatives are
ambiguous. Native pool-mass log-odds initialize the gate
(Equation~\ref{eq:gate-prior}), and zero terminal
residuals recover the native distribution renormalized within the pool.

Candidate memory entries are precomputed, but only the chosen entry becomes
visible. In Figure~\ref{fig:architecture}, repairing C to D changes the next
prefix to $[\mathrm{A,B,D}]$. The target then verifies the resulting single
path using its ordinary greedy rule. Global reads run once, followed by
$H$ dependent decisions; the 35.06M-parameter selector does not repeat a
full-vocabulary projection. Its benefit must cover both static read cost
and sequential selection cost, which we separate in the evaluation.

\section{Joint Training for Support and Repair}
\label{sec:training}
A selector cannot recover a token that its pool omits, and a strong pool
alone does not determine which choice will extend the accepted prefix.
DRelay therefore trains the backbone and selector jointly, while freezing
the target, token embedding, and vocabulary readout. During training,
prefix memory contains earlier target-greedy tokens under a strict $j<i$
mask. Their identities are fixed; gradients pass through draft features
and memory projections. Teacher proposal tokens never enter Global Read.

\paragraph{Learn support and selection together.}
For target label $y_i$, define coverage $c_i=\ind[y_i\in\mathcal V_i]$ and
its pool index $k_i$ when covered. A covered base-correct token teaches KEEP;
a covered alternative teaches both REPAIR and its identity:
\begin{equation}
 \ell_i^{\mathrm{sel}}=
 \begin{cases}
 \operatorname{BCE}(g_i,0),&y_i=v_{i0},\\
 \operatorname{BCE}(g_i,1)-\log r_{ik_i},&c_i=1,\ k_i>0,\\
 0,&c_i=0.
 \end{cases}
 \label{eq:selector-loss}
\end{equation}
BCE takes the gate logit as its first argument. A pool miss supplies no
valid selector action. Dense vocabulary supervision
still teaches the backbone to include the missing token. Combining both
responsibilities gives
\begin{equation}
 \mathcal L_t=\frac{\sum_{b,i}m_{b,i}
 [-\log q_{b,i}(y_{b,i})+\alpha_t w_{b,i}\ell_{b,i}^{\mathrm{sel}}]}
 {\sum_{b,i}m_{b,i}},
 \label{eq:loss}
\end{equation}
where $b$ indexes anchor blocks and $m$ excludes anchors, padding, and
noncontiguous response positions. The common valid-token denominator keeps
support and action learning on the same scale. Misses are neither KEEP
labels nor a separate output class.

\paragraph{Value a decision by the prefix it enables.}
The one-edit diagnostic shows why equal token credit can overlook a useful
repair: a decision matters only if decoding reaches it, and its value grows
with the continuation it enables. Let $p_i=p_i^{\mathrm{act}}(k_i)$ when covered and zero
otherwise, and set $\bar p_i=c_i(1+\operatorname{stopgrad}(p_i))/2$.
We assign raw credit
\begin{equation}
 v_i=c_i\underbrace{\prod_{j<i}\bar p_j}_{\text{reach}}
 \underbrace{\left(1+\sum_{t=i+1}^{H}\prod_{j=i+1}^{t}\bar p_j\right)}_{
 \text{continuation value}}.
 \label{eq:credit}
\end{equation}
These factors have the form of marginal prefix value, excluding the current
position's own probability. Within each valid block, we normalize credit
to mean one, retain a uniform floor, and bound its largest weight by four.
The resulting $w_i$ gradually replaces uniform weighting. Detachment and
bounded normalization prevent uncertain long products from dominating
learning or eliminating downstream supervision. This is a weighted
likelihood surrogate, not an unbiased gradient of greedy accepted length.

\paragraph{Learn at boundaries the current drafter reaches.}
Uniformly sampled anchors provide broad coverage but need not resemble
verification boundaries. We mix them with boundaries traced by a
no-gradient teacher-prefix scout. Before its first failure, the scout and
incremental native selection see the same earlier tokens; the scout can
therefore advance by $A+1$, where $A$ is the accepted proposal count.
Sampled anchors receive fresh differentiable forwards. Dense supervision
continues under teacher prefixes after a rejection, rather than treating
those suffixes as visited native states. The reference schedule increases
the native-anchor fraction from 30\% to 50\% and the selector coefficient
from approximately 0.05 to 1; bounded credit activates progressively.
The implementation details specify the exact schedule and information boundaries.

\section{Experiments}
\label{sec:experiments}
\subsection{Experimental setup}
\paragraph{Models and benchmarks.}
We evaluate Qwen3-4B and Qwen3-8B at temperatures 0 and 1 on eight datasets:
GSM8K~\citep{cobbe2021gsm8k}, MATH-500~\citep{lightman2023verify}, and
AIME25~\citep{opencompass2025aime} for mathematics; HumanEval~\citep{chen2021humaneval},
MBPP~\citep{austin2021program}, and LiveCodeBench~\citep{jain2024livecodebench}
for code; and MT-Bench~\citep{zheng2023mtbench} and Alpaca~\citep{taori2023alpaca}
for dialogue and instruction following. All eight datasets are evaluated under
the same decoding and hardware protocol. We compare DRelay with DFlash,
Domino, and DSpark, using target-only autoregressive decoding as the reference;
each method uses its native drafting and scheduling procedure.

\paragraph{Training and metrics.}
We train on the complete 1.42M PerfectBlend dataset~\citep{xu2024perfectblend} through SpecForge~\citep{li2026specforge} using eight
H800 GPUs and run inference on one H800. DRelay uses $H=15$ proposals and
$K=16$ candidates. We report end-to-end speedup $S$, including prefill and
all decoding overhead, and committed tokens per verification round $\tau$,
including the target bonus or replacement token. Unless stated otherwise,
all comparisons use the same target model, decoding budget, hardware, and
evaluation protocol.

\begin{table}[t]
\centering\scriptsize
\caption{\textbf{Single-H800 system comparison on Qwen3-4B and Qwen3-8B.} Each temperature reports speedup $S$ over AR with the \emph{same target and temperature}, and committed tokens per round $\tau$. LCB denotes LiveCodeBench. The reported average is the arithmetic mean over the eight listed datasets.}
\label{tab:main_results}
\setlength{\tabcolsep}{1.0pt}
\renewcommand{\arraystretch}{0.94}
\resizebox{\linewidth}{!}{%
\begin{tabular}{@{}l@{\hspace{4pt}}cc@{\hspace{4pt}}cc@{\hspace{4pt}}cc@{\hspace{4pt}}cc@{\hspace{4pt}}cc@{\hspace{4pt}}cc@{\hspace{4pt}}cc@{\hspace{4pt}}cc@{\hspace{4pt}}cc@{}}
\toprule
 & \multicolumn{6}{c}{\textsc{Math}} & \multicolumn{6}{c}{\textsc{Code}} & \multicolumn{4}{c}{\textsc{Chat}} & \multicolumn{2}{c}{\textsc{Overall}}\\
\cmidrule(lr){2-7}\cmidrule(lr){8-13}\cmidrule(lr){14-17}\cmidrule(l){18-19}
Method & \multicolumn{2}{c}{GSM8K} & \multicolumn{2}{c}{MATH-500} & \multicolumn{2}{c}{AIME25} & \multicolumn{2}{c}{HumanEval} & \multicolumn{2}{c}{MBPP} & \multicolumn{2}{c}{LCB} & \multicolumn{2}{c}{MT-Bench} & \multicolumn{2}{c}{Alpaca} & \multicolumn{2}{c}{Avg.}\\
\cmidrule(lr){2-3}\cmidrule(lr){4-5}\cmidrule(lr){6-7}\cmidrule(lr){8-9}\cmidrule(lr){10-11}\cmidrule(lr){12-13}\cmidrule(lr){14-15}\cmidrule(lr){16-17}\cmidrule(lr){18-19}
 & \multicolumn{1}{c}{$S$} & \multicolumn{1}{c}{$\tau$} & \multicolumn{1}{c}{$S$} & \multicolumn{1}{c}{$\tau$} & \multicolumn{1}{c}{$S$} & \multicolumn{1}{c}{$\tau$} & \multicolumn{1}{c}{$S$} & \multicolumn{1}{c}{$\tau$} & \multicolumn{1}{c}{$S$} & \multicolumn{1}{c}{$\tau$} & \multicolumn{1}{c}{$S$} & \multicolumn{1}{c}{$\tau$} & \multicolumn{1}{c}{$S$} & \multicolumn{1}{c}{$\tau$} & \multicolumn{1}{c}{$S$} & \multicolumn{1}{c}{$\tau$} & \multicolumn{1}{c}{$S$} & \multicolumn{1}{c}{$\tau$}\\
\midrule
\multicolumn{19}{c}{\textbf{Temperature = 0} (greedy)}\\
\midrule
\multicolumn{19}{l}{\textit{Qwen3-4B}}\\
DFlash & 3.05$\times$ & 6.40 & 3.99$\times$ & 7.89 & 3.60$\times$ & 7.32 & 3.24$\times$ & 6.61 & 3.47$\times$ & 6.01 & 3.76$\times$ & 7.40 & 1.96$\times$ & 4.11 & 1.83$\times$ & 3.48 & 3.11$\times$ & 6.15\\
Domino & 3.98$\times$ & 9.21 & 4.21$\times$ & 9.04 & 3.33$\times$ & 7.25 & 3.31$\times$ & 6.92 & 3.62$\times$ & 6.68 & 3.41$\times$ & 6.89 & 2.23$\times$ & 4.94 & 2.04$\times$ & 4.30 & 3.27$\times$ & 6.90\\
DSpark & 3.94$\times$ & 9.24 & 4.25$\times$ & 9.17 & 3.48$\times$ & 7.47 & 3.37$\times$ & 6.94 & 3.64$\times$ & 6.65 & 3.49$\times$ & 7.07 & 2.24$\times$ & 5.01 & 2.01$\times$ & 4.27 & 3.30$\times$ & 6.98\\
\textbf{DRelay} & \textbf{4.18}$\times$ & \textbf{9.97} & \textbf{4.49}$\times$ & \textbf{9.62} & \textbf{3.85}$\times$ & \textbf{8.37} & \textbf{3.63}$\times$ & \textbf{7.62} & \textbf{3.84}$\times$ & \textbf{7.09} & \textbf{4.07}$\times$ & \textbf{8.23} & \textbf{2.35}$\times$ & \textbf{5.27} & \textbf{2.15}$\times$ & \textbf{4.49} & \textbf{3.57}$\times$ & \textbf{7.58}\\
\midrule
\multicolumn{19}{l}{\textit{Qwen3-8B}}\\
DFlash & 3.47$\times$ & 6.29 & 4.60$\times$ & 8.03 & 4.26$\times$ & 7.29 & 3.70$\times$ & 6.46 & 3.31$\times$ & 5.88 & 3.66$\times$ & 7.20 & 1.91$\times$ & 4.09 & 1.72$\times$ & 3.61 & 3.33$\times$ & 6.11\\
Domino & 4.72$\times$ & 9.48 & 4.73$\times$ & 9.39 & 4.15$\times$ & 7.57 & 3.87$\times$ & 7.18 & 3.52$\times$ & 6.80 & 3.40$\times$ & 7.13 & 2.16$\times$ & 5.13 & 2.05$\times$ & 4.34 & 3.58$\times$ & 7.13\\
DSpark & 4.64$\times$ & 9.39 & 4.69$\times$ & 9.31 & 4.18$\times$ & 7.65 & 3.83$\times$ & 7.24 & 3.41$\times$ & 6.69 & 3.46$\times$ & 7.23 & 2.25$\times$ & 5.26 & 1.99$\times$ & 4.29 & 3.56$\times$ & 7.13\\
\textbf{DRelay} & \textbf{4.95}$\times$ & \textbf{10.03} & \textbf{5.01}$\times$ & \textbf{9.96} & \textbf{4.72}$\times$ & \textbf{8.61} & \textbf{4.20}$\times$ & \textbf{7.84} & \textbf{3.71}$\times$ & \textbf{7.18} & \textbf{4.01}$\times$ & \textbf{8.47} & \textbf{2.36}$\times$ & \textbf{5.39} & \textbf{2.14}$\times$ & \textbf{4.51} & \textbf{3.89}$\times$ & \textbf{7.75}\\
\midrule
\multicolumn{19}{c}{\textbf{Temperature = 1} (sampling)}\\
\midrule
\multicolumn{19}{l}{\textit{Qwen3-4B}}\\
DFlash & 2.84$\times$ & 5.93 & 3.09$\times$ & 6.71 & 2.36$\times$ & 4.71 & 2.98$\times$ & 6.01 & 2.66$\times$ & 5.56 & 2.94$\times$ & 7.14 & 1.58$\times$ & 3.92 & 1.44$\times$ & 3.37 & 2.49$\times$ & 5.42\\
Domino & 3.36$\times$ & 8.17 & 3.34$\times$ & 7.36 & 2.17$\times$ & 4.74 & 2.96$\times$ & 6.27 & 2.75$\times$ & 6.22 & 2.63$\times$ & 6.73 & 1.72$\times$ & 4.66 & 1.64$\times$ & 4.14 & 2.57$\times$ & 6.04\\
DSpark & 3.43$\times$ & 8.30 & 3.38$\times$ & 7.39 & 2.28$\times$ & 4.82 & 2.93$\times$ & 6.18 & 2.77$\times$ & 6.26 & 2.76$\times$ & 6.89 & 1.69$\times$ & 4.61 & 1.59$\times$ & 4.07 & 2.60$\times$ & 6.07\\
\textbf{DRelay} & \textbf{3.63}$\times$ & \textbf{8.86} & \textbf{3.54}$\times$ & \textbf{7.85} & \textbf{2.58}$\times$ & \textbf{5.56} & \textbf{3.14}$\times$ & \textbf{6.71} & \textbf{2.90}$\times$ & \textbf{6.57} & \textbf{3.17}$\times$ & \textbf{8.16} & \textbf{1.81}$\times$ & \textbf{4.78} & \textbf{1.68}$\times$ & \textbf{4.25} & \textbf{2.81}$\times$ & \textbf{6.59}\\
\midrule
\multicolumn{19}{l}{\textit{Qwen3-8B}}\\
DFlash & 3.17$\times$ & 5.86 & 3.44$\times$ & 6.57 & 2.71$\times$ & 4.80 & 3.14$\times$ & 5.50 & 2.91$\times$ & 5.25 & 3.52$\times$ & 7.17 & 1.70$\times$ & 3.67 & 1.61$\times$ & 3.48 & 2.78$\times$ & 5.29\\
Domino & 3.88$\times$ & 7.90 & 3.24$\times$ & 7.02 & 2.47$\times$ & 4.67 & 3.10$\times$ & 5.73 & 3.02$\times$ & 5.87 & 3.08$\times$ & 6.87 & 1.86$\times$ & 4.53 & 1.83$\times$ & 4.00 & 2.81$\times$ & 5.82\\
DSpark & 3.81$\times$ & 7.78 & 3.17$\times$ & 6.94 & 2.51$\times$ & 4.75 & 3.16$\times$ & 5.85 & 2.99$\times$ & 5.78 & 3.14$\times$ & 6.90 & 1.91$\times$ & 4.62 & 1.72$\times$ & 3.87 & 2.80$\times$ & 5.81\\
\textbf{DRelay} & \textbf{4.16}$\times$ & \textbf{8.51} & \textbf{3.49}$\times$ & \textbf{7.57} & \textbf{2.88}$\times$ & \textbf{5.43} & \textbf{3.41}$\times$ & \textbf{6.34} & \textbf{3.23}$\times$ & \textbf{6.15} & \textbf{3.64}$\times$ & \textbf{8.13} & \textbf{1.95}$\times$ & \textbf{4.74} & \textbf{1.94}$\times$ & \textbf{4.22} & \textbf{3.09}$\times$ & \textbf{6.39}\\
\bottomrule
\end{tabular}%
}
\end{table}

\subsection{System results}
All DRelay and baseline results in this subsection are obtained with the
SGLang serving framework~\citep{zheng2024sglang}.

\paragraph{Single-request decoding.}
Table~\ref{tab:main_results} reports speedup and committed tokens per round
over all eight datasets. DRelay has the highest average in every panel,
from $2.81\times$ at Qwen3-4B/temperature 1 to $3.89\times$ at
Qwen3-8B/temperature 0; the corresponding $\tau$ values rise with it.
Each panel uses its matching AR reference and an eight-dataset arithmetic mean.

\begin{figure}[t]
\refstepcounter{table}\label{tab:concurrency}
\edef\concurrencytabnum{\thetable}
\refstepcounter{figure}\label{fig:latency}
\centering
\begin{minipage}[t]{0.44\linewidth}
\vspace{0pt}
\centering
\includegraphics[width=\linewidth]{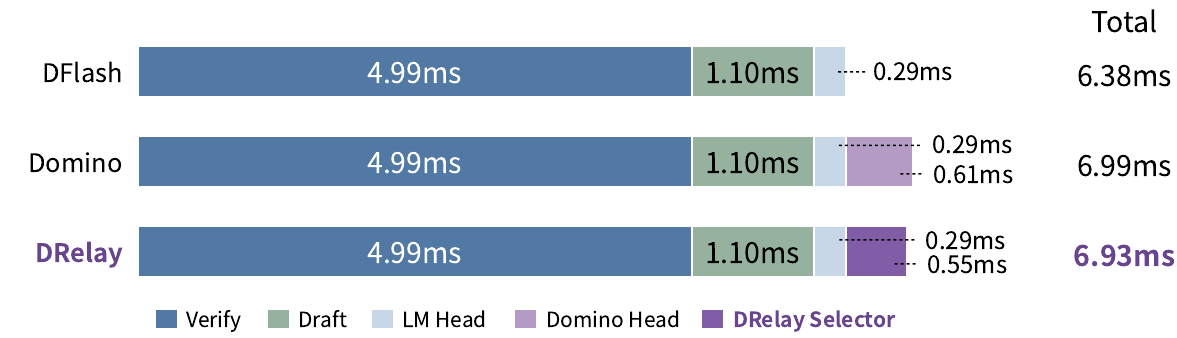}
\par\smallskip
\raggedright\footnotesize
\textbf{Figure~\thefigure:} Latency breakdown of DFlash, Domino, and
DRelay. Bars decompose latency into verification, drafting, LM-head
computation, and method-specific head or selector overhead.
\par\smallskip
\raggedright\footnotesize
Tables~\ref{tab:architecture-ablation} and~\ref{tab:training-ablation} compare
acceptance length $\tau$, using the same definition as Table~\ref{tab:main_results}; they isolate architectural and training-objective contributions.
\par\medskip
\centering
\refstepcounter{table}\label{tab:architecture-ablation}
\textbf{Table~\thetable: Architecture ablation.}
\par\smallskip
\footnotesize
\setlength{\tabcolsep}{3pt}
\renewcommand{\arraystretch}{1.18}
\begin{tabular*}{\linewidth}{@{}l@{\extracolsep{\fill}}c@{}}
\toprule
Variant & Accepted length\\
\midrule
Full & \textbf{7.58}\\
No Global Read & 7.16\\
\bottomrule
\end{tabular*}

\par\medskip
\refstepcounter{table}\label{tab:training-ablation}
\textbf{Table~\thetable: Training ablation.}
\par\smallskip
\footnotesize
\setlength{\tabcolsep}{3pt}
\renewcommand{\arraystretch}{1.18}
\begin{tabular*}{\linewidth}{@{}l@{\extracolsep{\fill}}c@{}}
\toprule
Variant & Accepted length\\
\midrule
Full & \textbf{7.58}\\
No support loss & 7.47\\
No selector loss & 7.22\\
Uniform repair weights & 7.19\\
\bottomrule
\end{tabular*}

\end{minipage}
\hfill
\begin{minipage}[t]{0.52\linewidth}
\vspace{0pt}
\centering
\textbf{Table~\concurrencytabnum: Concurrent serving.}
\par\smallskip
\raggedright\scriptsize
Throughput is in tokens/s; speculative rows include the speedup over AR at the
same concurrency as subscripts. Results use greedy decoding on AIME25 and
HumanEval.
\par\smallskip
\definecolor{speedupgreen}{RGB}{0,128,0}
\newcommand{\speedup}[1]{_{\textcolor{speedupgreen}{\text{#1}\times}}}
\scriptsize
\setlength{\tabcolsep}{1.8pt}
\renewcommand{\arraystretch}{0.96}
\begin{tabular}{@{}l@{\hspace{7pt}}ccccc@{}}
\toprule
 & \multicolumn{5}{c}{Concurrency}\\
\cmidrule(l){2-6}
Method & 2 & 4 & 8 & 16 & 32\\
\midrule
\multicolumn{6}{l}{\textit{Qwen3-4B (greedy)}}\\
\multicolumn{6}{l}{\textit{AIME25}}\\
AR & \(\text{433}\) & \(\text{811}\) & \(\text{1,415}\) & \(\text{2,490}\) & \(\text{4,310}\)\\
DFlash & \(\text{1,455}\speedup{3.36}\) & \(\text{2,757}\speedup{3.40}\) & \(\text{4,780}\speedup{3.38}\) & \(\text{6,638}\speedup{2.67}\) & \(\text{9,290}\speedup{2.16}\)\\
Domino & \(\text{1,423}\speedup{3.29}\) & \(\text{2,477}\speedup{3.06}\) & \(\text{4,299}\speedup{3.04}\) & \(\text{6,563}\speedup{2.64}\) & \(\text{9,006}\speedup{2.09}\)\\
DSpark & \(\text{1,447}\speedup{3.34}\) & \(\text{2,524}\speedup{3.11}\) & \(\text{4,418}\speedup{3.12}\) & \(\text{6,596}\speedup{2.65}\) & \(\text{9,104}\speedup{2.11}\)\\
\textbf{DRelay} & \(\text{\bfseries 1,614}\speedup{3.73}\) & \(\text{\bfseries 2,965}\speedup{3.66}\) & \(\text{\bfseries 4,914}\speedup{3.47}\) & \(\text{\bfseries 7,181}\speedup{2.88}\) & \(\text{\bfseries 10,077}\speedup{2.34}\)\\
\addlinespace
\multicolumn{6}{l}{\textit{HumanEval}}\\
AR & \(\text{442}\) & \(\text{853}\) & \(\text{1,571}\) & \(\text{2,771}\) & \(\text{3,880}\)\\
DFlash & \(\text{1,352}\speedup{3.06}\) & \(\text{2,425}\speedup{2.84}\) & \(\text{3,939}\speedup{2.51}\) & \(\text{6,089}\speedup{2.20}\) & \(\text{8,157}\speedup{2.10}\)\\
Domino & \(\text{1,419}\speedup{3.21}\) & \(\text{2,673}\speedup{3.13}\) & \(\text{4,190}\speedup{2.66}\) & \(\text{6,322}\speedup{2.28}\) & \(\text{8,483}\speedup{2.18}\)\\
DSpark & \(\text{1,443}\speedup{3.26}\) & \(\text{2,712}\speedup{3.18}\) & \(\text{4,217}\speedup{2.68}\) & \(\text{6,390}\speedup{2.30}\) & \(\text{8,511}\speedup{2.19}\)\\
\textbf{DRelay} & \(\text{\bfseries 1,484}\speedup{3.36}\) & \(\text{\bfseries 2,790}\speedup{3.27}\) & \(\text{\bfseries 4,362}\speedup{2.78}\) & \(\text{\bfseries 6,586}\speedup{2.38}\) & \(\text{\bfseries 8,805}\speedup{2.27}\)\\
\midrule
\multicolumn{6}{l}{\textit{Qwen3-8B (greedy)}}\\
\multicolumn{6}{l}{\textit{AIME25}}\\
AR & \(\text{282}\) & \(\text{532}\) & \(\text{943}\) & \(\text{1,636}\) & \(\text{3,120}\)\\
DFlash & \(\text{1,162}\speedup{4.11}\) & \(\text{2,020}\speedup{3.80}\) & \(\text{3,591}\speedup{3.81}\) & \(\text{5,082}\speedup{3.11}\) & \(\text{7,630}\speedup{2.45}\)\\
Domino & \(\text{1,092}\speedup{3.87}\) & \(\text{1,957}\speedup{3.68}\) & \(\text{3,344}\speedup{3.55}\) & \(\text{4,885}\speedup{2.99}\) & \(\text{7,351}\speedup{2.36}\)\\
DSpark & \(\text{1,114}\speedup{3.95}\) & \(\text{2,027}\speedup{3.81}\) & \(\text{3,502}\speedup{3.71}\) & \(\text{4,972}\speedup{3.04}\) & \(\text{7,428}\speedup{2.38}\)\\
\textbf{DRelay} & \(\text{\bfseries 1,226}\speedup{4.35}\) & \(\text{\bfseries 2,331}\speedup{4.38}\) & \(\text{\bfseries 3,807}\speedup{4.04}\) & \(\text{\bfseries 5,365}\speedup{3.28}\) & \(\text{\bfseries 8,203}\speedup{2.63}\)\\
\addlinespace
\multicolumn{6}{l}{\textit{HumanEval}}\\
AR & \(\text{292}\) & \(\text{566}\) & \(\text{1,074}\) & \(\text{1,955}\) & \(\text{2,582}\)\\
DFlash & \(\text{1,047}\speedup{3.59}\) & \(\text{1,914}\speedup{3.38}\) & \(\text{3,300}\speedup{3.07}\) & \(\text{5,088}\speedup{2.60}\) & \(\text{6,797}\speedup{2.63}\)\\
Domino & \(\text{1,084}\speedup{3.72}\) & \(\text{2,008}\speedup{3.55}\) & \(\text{3,429}\speedup{3.19}\) & \(\text{5,295}\speedup{2.71}\) & \(\text{7,120}\speedup{2.76}\)\\
DSpark & \(\text{1,108}\speedup{3.79}\) & \(\text{2,087}\speedup{3.69}\) & \(\text{3,501}\speedup{3.26}\) & \(\text{5,386}\speedup{2.75}\) & \(\text{7,231}\speedup{2.80}\)\\
\textbf{DRelay} & \(\text{\bfseries 1,153}\speedup{3.95}\) & \(\text{\bfseries 2,094}\speedup{3.70}\) & \(\text{\bfseries 3,566}\speedup{3.32}\) & \(\text{\bfseries 5,497}\speedup{2.81}\) & \(\text{\bfseries 7,464}\speedup{2.89}\)\\
\bottomrule
\end{tabular}

\end{minipage}
\end{figure}

\paragraph{Concurrent serving.}
Table~\ref{tab:concurrency} reports throughput on AIME25 and HumanEval for
concurrency 2--32. DRelay is fastest at every setting; at concurrency 32 its
speedups over AR are $2.34\times$/$2.27\times$ on Qwen3-4B and
$2.63\times$/$2.89\times$ on Qwen3-8B (AIME25/HumanEval). The lead persists
as speedup over AR decreases with load. Baseline rows are tokens/s;
speculative rows add same-concurrency speedups as subscripts.

\paragraph{Efficiency.}
Figure~\ref{fig:latency} decomposes per-round latency into verification,
drafting, language-model head computation, and method-specific overhead.
DRelay adds 0.55\,ms of selector computation to the 6.38\,ms shared by
verification, drafting, and the language-model head, yielding a total of
6.93\,ms. This is 8.6\% above DFlash's 6.38\,ms and slightly below Domino's
6.99\,ms. Verification remains the largest component at 4.99\,ms. The
end-to-end results in Tables~\ref{tab:main_results} and
\ref{tab:concurrency} show that the longer verified prefixes compensate for
this additional per-round cost.

\subsection{Ablation studies}
Both tables report mean accepted length $\tau$ over the same eight datasets
used in Table~\ref{tab:main_results}, using the same target/temperature
setting as Full and the same hardware and decoding protocol. Each ablation changes only the
component named in its row; all other components, positions, data, and training
settings remain identical.

\paragraph{Global candidate context.}
Table~\ref{tab:architecture-ablation} compares the full model with No Global Read,
which removes global slot mixing and candidate reads. The accepted length falls
from $7.58$ to $7.16$, a reduction of $0.42$ tokens (5.5\%). This drop shows
that cross-position candidate evidence contributes beyond the remaining local
and prefix information.

\paragraph{Training objective.}
Table~\ref{tab:training-ablation} compares the full objective with three
single-component controls. Removing the support loss reduces accepted length
to $7.47$ ($-0.11$), removing selector supervision reduces it to $7.22$
($-0.36$), and replacing prefix-value weights with uniform weights reduces it
to $7.19$ ($-0.39$). The larger degradation from removing selector supervision
or prefix-value weighting indicates that learning when to repair and assigning
more credit to prefix-enabling repairs are both important. The support-loss
control retains selector supervision, the selector-loss control retains the
draft objective but disables repair, and the uniform-weight control retains both
losses; all remaining settings are unchanged.

\section{Conclusion}
DRelay treats a parallel draft as evidence for its own token choices.
Native-draft diagnostics connect three opportunities: early mistakes often
retain the correct candidate, reliable later predictions help distinguish
that candidate, and a precise repair can recover a consecutive continuation.
The architecture follows this logic by reading candidate-level global
context once, then making selective repairs against the actual selected
prefix. Joint support and repair learning aligns these decisions with
prefix reach and continuation value. The system results quantify the end-to-end trade-off, while the latency
breakdown identifies the cost of the selector relative to verification and
drafting.

\label{mainend}
\section*{AI Use Statement}
AI was used only for literature retrieval and discovery (e.g., identifying related work) and for assistance with LaTeX formatting and typesetting. All research ideas, methodological and experimental design, analyses, figures, scientific claims, and the manuscript text were independently developed and written by the authors.
\bibliographystyle{iclr2027_conference}
\bibliography{references}
\clearpage
\appendix
\section{Reference Diagnostics}
\label{app:diagnostics}

This appendix defines the offline diagnostics used to motivate DRelay in
Section~\ref{sec:motivation}.  They are computed from frozen decoding traces
and are separate from the trained-system evaluation in
Section~\ref{sec:experiments}.  In particular, the diagnostics measure whether
useful evidence is present in a parallel draft; they do not use oracle labels
as inputs to DRelay at inference time.

\paragraph{Reference traces and terminology.}
We collect native parallel-drafting traces from DFlash and Domino
\citep{chen2026dflash,huang2026domino} with Qwen3-4B as the target model
\citep{yang2025qwen3}.  At each position, the \emph{base} token is the
backbone argmax and the \emph{served} token is the token selected by the
corresponding native drafting procedure.  The target's greedy continuation
under the verified prefix supplies the reference token used to identify a
mismatch.  The original top-$16$ candidate pool is retained for the
candidate-availability analyses.  All traces use the same tokenizer and
verification rule as the corresponding main-text diagnostic.

\paragraph{Candidate coverage.}
For a position $i$, let $y_i$ be the target reference token and let
$\mathcal V_i$ be the original candidate pool.  We record
$c_i=\mathbb 1[y_i\in\mathcal V_i]$.  Coverage is evaluated at the first base
error and at the first error in the native served path, with the latter
restricted to positions whose served pool is defined.  These strata localize
where a repair is possible; they do not imply that a deployed selector can
observe the target token or the error location.

\paragraph{Continuation evidence.}
For a candidate $v\in\mathcal V_i$, we score a later predicted continuation
under the verified prefix extended by $v$.  The score uses the target model's
conditional likelihood of the predicted tokens after position $i$ and omits
the current-token target probability.  Replacing the continuation by a
correctness-matched continuation from another prompt provides the control
comparison in Figure~\ref{fig:motivation}.  The control changes the future
content while preserving the task and reliability strata, so a positive
real-minus-control contrast indicates context-aligned evidence rather than
candidate prior alone.

\paragraph{One-edit repair value.}
Let $A$ be the number of proposals accepted before the first served error and
let $R$ be the number of immediately following proposals that agree with the
target.  If the target token is covered, changing only that error increases
the accepted prefix by $1+R$; later matches separated by another error do not
count.  We report this quantity as an oracle opportunity analysis over
complete blocks.  It is not an estimate of DRelay's realized gain, because
DRelay must infer both whether to edit and which candidate to select.

\paragraph{Interpretation and scope.}
The diagnostics establish the existence of candidate availability and
continuation evidence that a compact selector may exploit.  They do not
provide supervision from unverified target states, do not evaluate a learned
repair policy, and do not replace the end-to-end speed and acceptance results
reported in Section~\ref{sec:experiments}.

\FloatBarrier
\section{Implementation and Training Details}
\label{app:design}

This appendix records the implementation contract shared by the reported
experiments.  The equations in Sections~\ref{sec:method} and
\ref{sec:training} give the full computation; the points below make explicit
the information boundaries and supervision used in the reported configuration.

\paragraph{Information boundary.}
At each round the parallel drafter produces hidden representations and a
candidate distribution for each of $H=15$ positions.  DRelay retains the
native top-$K$ pool with $K=16$.  Global Read consumes only these draft-side
representations and candidate distributions.  It never reads a teacher
proposal token or a future token that has not been verified by the target
model.  The Causal Selector receives the resulting candidate-specific global
features, the strictly verified history, and a prefix memory whose entry at
position $j$ is visible only to decisions at positions $i>j$.  Once a
candidate is selected, it is the only token written into the prefix used by
later decisions in the same block.

\paragraph{Trainable and frozen components.}
The target model, token embedding, and vocabulary readout remain frozen.  The
parallel drafting backbone, Global Read, and Causal Selector are trained
jointly.  The selector has separate KEEP/REPAIR and ranking branches.  Its gate starts
from the native pool-mass log-odds, with a learned residual correction:
\begin{equation}
 g_i=\log\sum_{k>0}q_i(v_{ik})-\log q_i(v_{i0})+\Delta g_i.
 \label{eq:gate-prior}
\end{equation}
Here $v_{i0}$ is the native candidate and $q_i$ is the draft distribution over
the retained pool.  Residual corrections are learned from the draft-side
features and verified history.  This preserves the native choice when
contextual evidence is insufficient and allows a repair to be carried forward
causally.

\paragraph{Supervision.}
For a target token $y_i$, the dense support term trains the draft distribution
to place mass on $y_i$, including when the token is absent from the retained
pool.  The selector term labels a covered base-correct token as KEEP and a
covered alternative as REPAIR with its target-pool index.  A pool miss has no
valid selector action and therefore contributes only to support supervision.
The selector term is multiplied by the prefix-value weights defined in
Equation~\ref{eq:credit}; these weights favor decisions that are reachable from
the current prefix and that preserve a correct continuation.  Padding,
anchors, and non-contiguous response positions are masked from both terms.
No future target token is supplied to Global Read or to the selector as an
input.

\paragraph{Training and serving protocol.}
We train on the complete PerfectBlend corpus through SpecForge, using the
training and hardware protocol described in Section~\ref{sec:experiments}.
All comparisons use the same target model, candidate budget, decoding budget,
and evaluation hardware.  During serving, Global Read is computed once per
parallel draft and is followed by the $H$ causal selector decisions; the target
then verifies the resulting single path and supplies the next anchor.  Thus the
reported latency includes both the shared draft/verification work and the
method-specific selection overhead shown in Figure~\ref{fig:latency}.

\FloatBarrier
\section{Diagnostic Scope and Interpretation}
\label{app:global-signal}

The motivation analysis asks whether later parallel predictions contain
information that can distinguish candidates at an earlier position.  This
appendix clarifies the diagnostic boundary used for that question.

\paragraph{What is scored.}
For a candidate $v$ at position $i$, the offline assessor conditions on the
verified prefix followed by $v$ and scores only the later predicted draft
tokens.  The current target-token probability is excluded, so the score
measures continuation compatibility rather than simply reproducing the native
candidate prior.  Real continuations are compared with correctness-matched
continuations from other prompts.  The target model is used as an offline
assessor for this measurement; it is not queried for unverified future states
by DRelay's Global Read.

\paragraph{What the diagnostic does not claim.}
First-error localization and correctness filters are used only to form
interpretable offline strata.  They are not available to the deployed
selector, and they do not define an inference-time repair rule.  Likewise, an
oracle one-edit gain measures the value of a possible correction under the
original candidate pool; it is not an end-to-end speedup or an accepted-length
prediction for the trained system.

\paragraph{Connection to DRelay.}
The diagnostic supports the design choice of reading global candidate context
while retaining a causal prefix selector.  Section~\ref{sec:method} implements
that boundary using draft-side hidden states, candidate distributions, and
verified history, and Section~\ref{sec:experiments} measures the resulting
trained system on all eight benchmarks.  The appendix therefore serves as a
definition of the evidence used in the motivation, rather than as a second
set of model or benchmark results.

\FloatBarrier
\end{document}